\documentclass{article}

\usepackage{arxiv}

\usepackage[utf8]{inputenc} % allow utf-8 input
\usepackage[T1]{fontenc}    % use 8-bit T1 fonts
\usepackage{hyperref}       % hyperlinks
\usepackage{url}            % simple URL typesetting
\usepackage{booktabs}       % professional-quality tables
\usepackage{amsfonts}       % blackboard math symbols
\usepackage{nicefrac}       % compact symbols for 1/2, etc.
\usepackage{microtype}      % microtypography
\usepackage{lipsum}
\usepackage{graphicx}
\title{Photofeeler-D3: A Neural Network with Voter Modeling for Dating Photo Impression Prediction}

\author{
  Agastya Kalra \thanks{This author served as a research partner for the duration of this project but is not an employee at Photofeeler Inc. For any inquiries related to Photofeeler Inc. please email ben@photofeeler.com} \\
  Computer Vision Advisor \\
  Photofeeler Inc. \\
  Ottawa, ON \\
  \texttt{agastya.kalra@gmail.com} \\
   \And
  Ben Peterson \\
  Co-founder, CTO\\
  Photofeeler Inc.\\
  Denver, CO \\
  \texttt{ben@photofeeler.com} \\
}

\begin{document}
\maketitle

\begin{abstract}
In just a few years, online dating has become the dominant way that young people meet to date, making the deceptively error-prone task of picking good dating profile photos vital to a generation's ability to form romantic connections. Until now, artificial intelligence approaches to Dating Photo Impression Prediction (DPIP) have been very inaccurate, unadaptable to real-world application, and have only taken into account a subject's physical attractiveness. To that effect, we propose Photofeeler-D3 - the first convolutional neural network as accurate as 10 human votes for how smart, trustworthy, and attractive the subject appears in highly variable dating photos. Our "attractive" output is also applicable to Facial Beauty Prediction (FBP), making Photofeeler-D3 state-of-the-art for both DPIP and FBP. We achieve this by leveraging Photofeeler's Dating Dataset (PDD) with over 1 million images and tens of millions of votes, our  novel  technique  of  voter  modeling, and cutting-edge computer vision techniques.
\end{abstract}

% % keywords can be removed
% \keywords{Computer Vision \and Dating \and More}

\section{Introduction}
Online dating is the future. \cite{onlinedatingisthefuture} In 2017, Tinder, a mobile dating app, became the number one top grossing app in the Apple App Store. \cite{tindertopgrossingapp} In 2019, 69\% of Generation Z is active on dating apps, making online dating the dominant means of meeting people to date for 18-to-22-year-olds today. \cite{onlinedatinggenz}\cite{onlinedatinggenz2}

But Generation Z's record loneliness \cite{lonelinessgenz} may point to the ineffectiveness of the current dating platforms. The leading dating apps' profiles are highly dependent on photos. Research says that photos are misleading because different photos of the same person can give entirely different impressions. \cite{differentphotossameperson} To make matters worse, individuals display bad judgment in choosing their own photos. \cite{selfimageselection} But according to The Guardian, 90\% of people decide to date someone based on their dating photos alone \cite{datingphoto} - meaning that picking the right photo is vital to one’s success.  This is why assistance in choosing dating profile photos is sorely needed in order to facilitate the right connections. The lifelong partnerships of many millions of people depend on this.

In industry, besides hiring an expert \cite{vida}, there are two types of online services used to evaluate photos for dating profiles: online voting websites \cite{pfmain}, and online artificial intelligence platforms (OAIPs) \cite{prettyscale, hotness}. Photofeeler \cite{pfmain} is the largest of the first category, with over 90 million votes cast, 2 million images voted on, and over 100k new votes every day. In this work, we  show that the Photofeeler-D3 network achieves the same statistical significance as 10 (unnormalized and unweighted) human votes, making it more accurate than a small group of independent humans, but not as accurate as a Photofeeler voting test. In terms of the OAIPs, the most popular platforms are hotness.ai \cite{hotness} and prettyscale.com \cite{prettyscale}, both of which only measure attractiveness. They are evaluated on the London Faces Dataset \cite{londonface} in a 2018 study \cite{study}. We outperform both of these online services on that benchmark by achieving over 28\% higher correlation with human voters.

While optimizing for the most attractive photo is a good proxy for maximizing matches, attractiveness alone is not the optimal metric if the goal is to find high quality matches that lead to actual dates and long-term relationships \cite{pfdating2018}. That is why Photofeeler's voting-based online dating photo rating service also measures the smart and trustworthy traits. This allows users to find the photo that not only makes them look \emph{hot}, but also reliable, principled, intellectual, and safe to meet with in person. With this in mind, the Photofeeler-D3 neural network outputs scores for these 3 traits - the first neural network to do so.

In literature, the closest well-studied task is Facial Beauty Prediction (FBP) \cite{xie2015scut, xu2015new, xu2017facial, hotornot, zhai2019beautynet, xu2018transferring, xu2018crnet, gray2010predicting, anderson2018facial, liu2017facial}. In FBP, the goal is to take a perfectly cropped photo of the subject's face looking forward in a neutral position, and predict the objective attractiveness of that individual \cite{xie2015scut}. In our case, the photos are of people in different settings, poses, expressions, outfits, makeup, lighting, and angles, taken with a variety of cameras. We show that our model's attractiveness output also works for FBP, achieving state-of-the-art performance on the benchmark SCUT-FBP dataset \cite{xie2015scut}. 

FBP has received some backlash on social media \cite{reddit} due to the ethics of objectively assigning attractiveness scores to individuals. In DPIP, the ratings are assigned to the photos, not the individual. Figure 1 shows photos from the Photofeeler Dating Dataset (PDD) of the same person with very different scores. The goal of DPIP is to give people the best chance at successfully finding long-term relationships in dating apps through selecting photos for the profile as \emph{objectively} as possible. We discuss FBP methods further in section 2, and compare to existing benchmarks in section 4.

\begin{figure}[ht]
\vskip 0.2in
\begin{center}
\centerline{\includegraphics[width=340px]{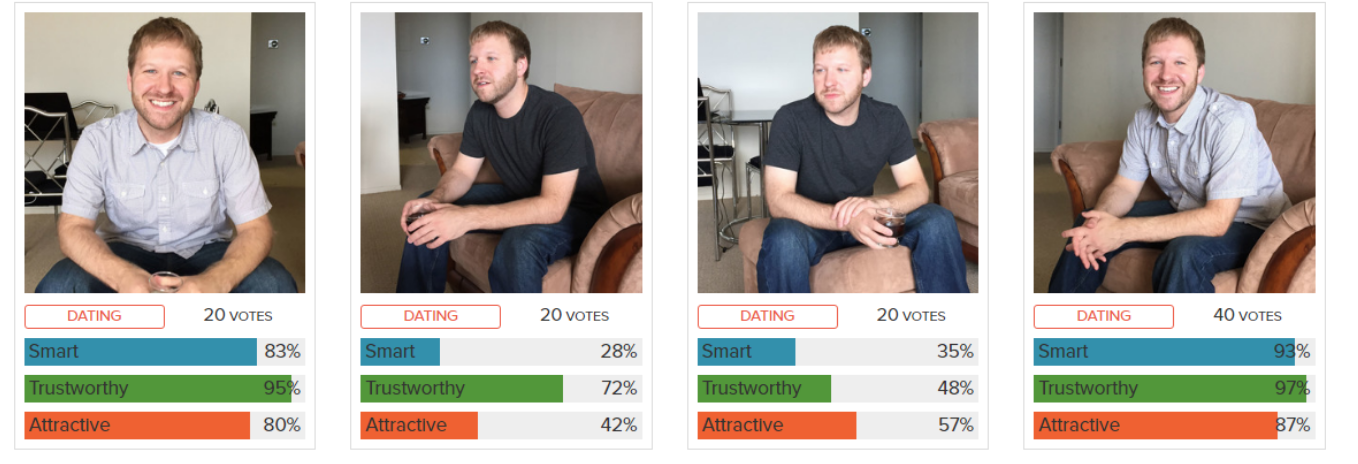}}
\caption{Four example photos of the same subject with different scores from the Photofeeler Dating Dataset.}
\label{subject_examples}
\end{center}
\vskip -0.2in
\end{figure} 

In this work, we explore the idea of using AI to predict the impressions given off by dating photos. We create a neural network that achieves state-of-the-art results on a variety of benchmark datasets \cite{londonface, xie2015scut, hotornot} and matches the accuracy of a small group of human voters for DPIP. We introduce voter modeling as an alternative solution to predicting average scores for each trait, which helps lower the impact of noise that comes from images without many votes. Finally we discuss the implications of our results on using votes to rate the smart, trustworthy, and attractive traits in single-subject photos.

The remainder of the paper is structured as follows. Section 2 reviews similar public datasets, convolutional neural networks, approaches for FBP, and online AI services for DPIP. Section 3 describes the PDD structure and the Photofeeler-D3 architecture and training procedure. Section 4 contains results on benchmark datasets and discussion. Section 5 summarizes the findings of the paper. 

\section{Related Work}
\label{sec:related work}

In this section we discuss the relevant benchmark datasets, convolutional neural network architectures, facial beauty prediction, and OAIPs.

\paragraph{Datasets}
There are a variety of benchmark datasets for scoring images: The AVA dataset \cite{ava}, the Hot-Or-Not dataset \cite{hotornot}, the SCUT-FBP dataset \cite{xie2015scut}, the LSFCB dataset \cite{zhai2019beautynet}, the London Faces Dataset \cite{londonface}, and the CelebA dataset \cite{celeba}. The AVA dataset \cite{ava} doesn't have attractiveness ratings for the subject, instead they have an attractiveness rating for the entire image i.e. \emph{Is this a good photo?}, which is very different from \emph{Does the subject look good in this photo?}. The Hot-Or-Not \cite{hotornot} dataset contains 2k images of single subject photos with at least 100 votes from the opposite sex on a 1-10 attractiveness scale. We report performance on this dataset since this is the closest publicly available dataset to our own. The SCUT-FBP \cite{xie2015scut} dataset is the standard benchmark for the FBP task - containing 500 images of cropped Asian female faces in neutral position staring forward into the camera. We benchmark our Photofeeler-D3 architecture on the SCUT-FBP dataset since the task is similar. The London Faces dataset is similar to the SCUT-FBP dataset except it contains 102 images of diverse males and females. It was used to benchmark prettyscale.com \cite{prettyscale} and hotness.ai \cite{hotness}, so we use it to benchmark our Photofeeler-D3 network. The LSFCB \cite{zhai2019beautynet} dataset contains 20k images for FBP but is not publicly available, so we do not include it. The CelebA \cite{celeba} dataset contains a binary indicator for attractiveness marked by a single labeler for each image, which is very different from DPIP, so we do not include it in our work. 

\begin{figure}[ht]
\vskip 0.2in
\begin{center}
\centerline{\includegraphics[width=400px]{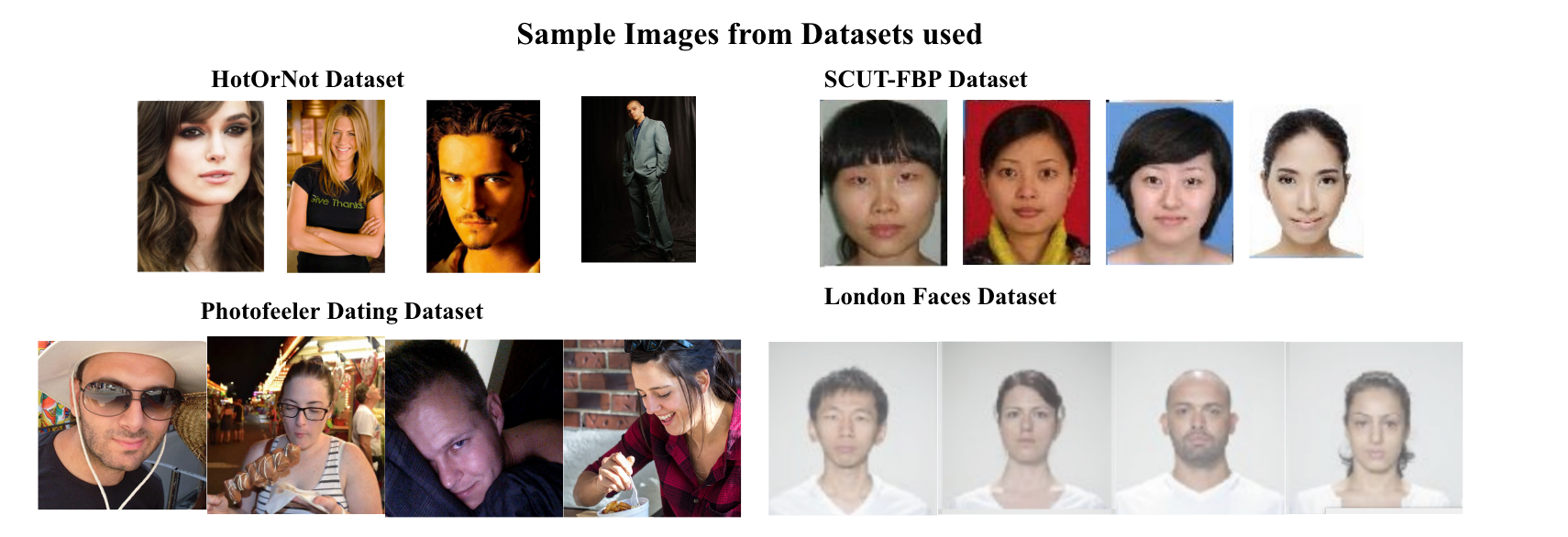}}
\caption{Sample photos from each dataset. The London Faces Dataset and the SCUT-FBP dataset are simpler than the HotOrNot dataset and the Photofeeler Dating Dataset. }
\label{dataset_examples}
\end{center}
\vskip -0.2in
\end{figure} 

\paragraph{Convolutional Neural Networks} 
In the last six years, convolutional neural networks (CNNs) have achieved state-of-the-art results in a variety of computer vision tasks including classification \cite{alexnet, inception3, inception4, resnet, vgg16, xception}, bounding box prediction \cite{fasterrcnn}, and image segmentation \cite{unet}. We present a brief review of relevant CNN architectures. \emph{Architectures:} The first major CNN architecture to be popularized was AlexNet \cite{alexnet} after its 2012 ILSVRC \cite{imagenet} win. It had 8 layers, used large convolution kernels and was the first successful application of dropout. After that, a variety of improvements have come along. VGG16 \cite{vgg16} won ILSVRC in 2014 by using many small kernels rather than a few large ones. 2015 was dominated by Residual Networks (ResNets) \cite{resnet} where they introduced the idea of deep architectures with skip connections. 2016 was won by the InceptionResNetV2 \cite{inception4}, which combined the inception architecture \cite{inception3} with skip connections to achieve even higher accuracy. In 2017 the Xception \cite{xception} architecture was introduced, which matched the performance of InceptionResNetV2 with much fewer parameters by leveraging depth-wise separable convolution layers. In 2018, the Neural Architecture Search Network \cite{nasnet} (NASNet) was published - an architecture generated through reinforcement learning. However, due it its size and complexity, it has yet to gain popularity. In our work we compare all architectures listed here since ResNet, not including NASNet. 

\paragraph{Facial Beauty Prediction} 
Facial Beauty Prediction is the task of objectively assessing the average attractiveness rating of a face in a neutral position looking forward into the camera \cite{xie2015scut}. This is very different from DPIP because in DPIP the subject is rated in different contexts. Traditional FBP algorithms \cite{xie2015scut} relied on facial landmarks and some combination of hand-engineered rules and shallow machine learning models. However since 2015, CNNs have dominated the FBP task \cite{xu2015new, xu2017facial, xu2018transferring, xu2018crnet, anderson2018facial, zhai2019beautynet, gray2010predicting, wang2014attractive} due to the wide availability of pretrained networks and increased access to public data. Gray et al. \cite{gray2010predicting} proposed a 4 layer CNN and were the first to discard facial landmarks. Gan et al. \cite{gan2014deep} used deep learning to extract beauty features instead of artificial feature selection. Xu et al. \cite{xu2015new} used a specific 6 layer CNN that took as input both the RGB image and a \emph{detail image} for facial beauty prediction on the SCUT-FBP \cite{xie2015scut} dataset. PI-CNN \cite{xu2017facial} - a psychology inspired convolutional neural network, introduced by Xu et al., separated the facial beauty representation learning and predictor training. Xu et al. \cite{xu2018transferring} proposed using models pretrained on other facial tasks as  a starting point to address the lack of data for FBP. Anderson et al. \cite{anderson2018facial} benchmark a variety of CNN architectures on the CelebA dataset for binary attractiveness prediction. Both Fan et al. \cite{fan2018label} and Liu et al. \cite{liu2017facial} propose replacing the regression output with a distribution prediction output and using a KL-Divergence loss rather than the standard mean squared error. We adopt a similar architecture to this. Gao et al. \cite{gaol2018automatic} utilize a multi-task learning training scheme where the model is required to output facial key-points along with average attractiveness scores. In CR-Net \cite{xu2018crnet}, Xu et al. propose using a weighted combination of mean squared error and cross-entropy loss to improve resilience to outliers when training. All of these works benchmark on either the HotOrNot \cite{hotornot} dataset or the SCUT-FBP \cite{xie2015scut} dataset. We benchmark Photofeeler-D3 on both.

\paragraph{Online AI Platforms}
There are 2 main OAIPs for attractiveness scoring: hotness.ai \cite{hotness} and prettyscale.com \cite{prettyscale}. Neither of these measure the smart or trustworthy traits. hotness.ai \cite{hotness} takes a photo of a single subject in any pose and, using an unspecified deep learning algorithm based on facial embedding, gives a discrete attractiveness score from 1-10. prettyscale.com \cite{prettyscale} requires the user to give a photo of a face and specify some geometric facial key-points. It uses a variety of hand-crafted rules to give an attractiveness score from 0-100. Photofeeler-D3 is different in that it takes in an image of a single subject in any pose and outputs smart, trustworthy, and attractive scores. Additionally, we conjecture that the Photofeeler-D3 network is the only one to use voter modeling. We show comparisons to both these tools in section 4.

\section{Our Method}
\label{sec:our method}

In this section we discuss the Photofeeler Dating Dataset and the Photofeeler-D3 neural network. 

\subsection{Photofeeler Dating Dataset}

The PDD contains ~1.2 million dating photos - ~1 million male images of ~200k unique male subjects and ~200k female images of ~50k unique female subjects. The images have a variety of aspect ratios, but the maximum side is at most 600 pixels. The metadata for each image contains a list of voters, a weight from $0-1$ for each vote (used to filter out poor quality votes), and both their normalized vote in the range $0-1$ and their original \emph{raw} vote in the range $0-3$ for each of the 3 traits. We normalize the votes for each voter depending on how they vote, i.e. if a voter gives mostly 0s and 1s, then a 2 from that voter will have a much higher normalized score than a voter who normally gives 2s and 3s. The weights are determined by how predictable a voter is, so a voter who always votes 1 will have a weight of 0. We exclude the weighting and normalization algorithms since they are Photofeeler Intellectual Property, however these algorithms dramatically improve the quality of the scores. All voters are the opposite sex of the subject in the photo. We compute the test labels $y_{it}$ for each image $x_i$ as a weighted sum of all the normalized votes $v_{ijt}$ where $i$ is the image index, $j$ is the voter index, $t$ is the trait (one of smart, attractive, or trustworthy) and $\Gamma_i$ is the set of voters that voted on the image $x_i$. It is important to note that these labels are not the "true score" of the image, as these traits are subjective. Rather they are noisy estimates of the population mean scores. We will demonstrate later how modeling this subjectivity is critical to our approach.

\begin{equation}
    y_{it} = \frac{\sum_{j \epsilon \Gamma_i}{w_j * v_{ijt}}}{\sum_{j \epsilon \Gamma_i}{w_j}}
\end{equation}

We separate out 10000 male subject images and 8000 female subject images to be used for testing. We guarantee that these images are of subjects not found in the training set and contain at least 10 votes to ensure some amount of statistical significance. We choose not to restrict it further since images with a higher number of votes tend to have higher scores because images getting low scores tend to be replaced more quickly by the user. Thus restricting it further would reduce the diversity of images in the test set. 

For experimentation on hyperparameters, we set aside what we call the \emph{small dataset}, a 25,311 male image subset of the training set (20,000 train, 3,000 val, 2,311 test), and evaluate only on the attractiveness class. Evaluating on male attractiveness will provide a lower bound for all classes because it is the most difficult to predict out of all traits and demographics. For training and evaluating our final set of hyperparameters, we use all male and female images in the PDD, we call this the \emph{large dataset}. When comparing the Photofeeler-D3 neural network to other benchmarks and to humans, we use the result of training on these 2 sets. We evaluate the model by using the Pearson correlation coefficient (PC) rather than mean squared error (MSE) because a model that simply predicts the sample mean of the training set for each test image would get a decent MSE, but would get a PC of 0\%.

The dataset was collected using Photofeeler's online voting platform \cite{pfmain}. Users upload their images and then receive votes based on how many votes they cast. The voters are given an image and told to rate their impression of each trait on a scale of 0 (No) to 3 (Very). These votes are weighted and normalized to create a score for each image on a scale of 0-1. Due to improvements in the normalization and weighting algorithm over time, the training images collected earlier have a higher error rate. The test set was taken from recent images only.

\subsection{Photofeeler-D3 Neural Network}
\subsubsection{Architecture}
\label{sec: architecture}
To effectively evaluate the impressions given off by images we use a 4 part architecture - the base network, the temporary output, the voter model, and the aggregator. We go through each component with respect to a single trait for simplicity. Extending to multiple traits just requires an output per trait. 
\begin{figure}[ht]
\vskip 0.2in
\begin{center}
\centerline{\includegraphics[width=340px]{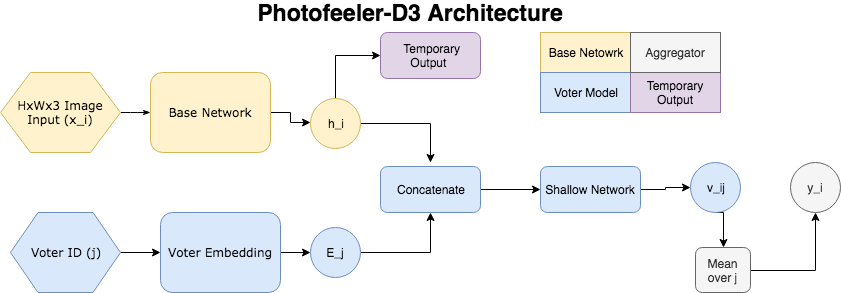}}
\caption{A diagram of the Photofeeler-D3 Architecture}
\label{Photofeeler-D3 Diagram}
\end{center}
\vskip -0.2in
\end{figure} 

\emph{Base Network:} We take as input an HxWx3 RGB image and forward pass it through one of the standard classification architectures \cite{inception4, xception, vgg16, resnet} mentioned in section 2.2. This is our base network $h_i = g(x_i; \theta)$, where $x_i$ represents the input image, the vector $h_i$ represents the output of the base network, and $\theta$ represents the parameters of the model. The choice of architecture and image input size are hyperparameters. In our best model we used the Xception \cite{xception} architecture with input image size 600x600x3. We center the images and pad the rest with black for non-square images. To get the vector $h_i$, we remove all fully connected layers from the architecture and apply global average pooling to the last convolutional layer.  
\begin{equation}
    h_i = g(x_i; \theta)
\end{equation}

\emph{Temporary Output:} Starting with $h_i$, we apply a fully connected layer with weight matrix $W_t$ for each trait $t$ and produce a per-trait output of $\bar{o}_{ti}$. The output is a 10-dimensional vector that has gone through the softmax $\sigma$ activation function. This output is used during training but removed during testing. Note we omit the $t$ subscript from the equation for simplicity. 
\begin{equation}
    \bar{o}_{i} = \sigma(Wh_i)
\end{equation}

\emph{Voter Model:} We introduce an embedding matrix $E$ that contains one row for each unique voter. To compute the predicted normalized vote score $\bar{v}_{ij}$ for voter $j$ on image $x_i$, we concatenate the output of the base network $h_i$ with the voter embedding $E_j$ and apply a shallow fully connected network $\phi$ that produces a distribution over 10 classes. For the label, we turn $v_{ij}$ into a 10-dimensional 1-hot encoding by rounding $v_{ij}$ to the nearest 0.1. To construct the real number $\bar{v}_{ij}$, we take the inner product of the output of $\phi$ with the vector $b$. The vector $b$ is defined as $[0.05, 0.15, 0.25 ... 0.95]$.
\begin{equation}
    \bar{v}_{ij} = <\phi([h_i, E_j]), b>
\end{equation}

\emph{Aggregator:} To produce the final prediction $\bar{y}_{i}$ we aggregate the predicted vote of a random sample $\beta$ of 200 voters. This allows us to get a stable estimate of what many voters will think of a single image. 

\begin{equation}
    \bar{y}_{i} = \frac{1}{n} \sum_{j \epsilon \beta}\bar{v}_{ij}
\end{equation}

Voter modeling means that the model is no longer trying to predict a potentially noisy estimate of the population mean. Instead it is predicting the subjective opinion of a particular voter - the score they would give to the image. This means the number of votes per image is less relevant to the training process and therefore the impact of noise that comes from images without many votes is limited. Then we take the mean of many predicted votes in the aggregation step to get a stable prediction of the scores of an image. We demonstrate the effectiveness of this system in section 4.  

\subsubsection{Training}
A separate model is trained for images with female subjects and images with male subjects. We follow the same 2-part training scheme for both; the first part trains the base network and the second part trains the voter model. We split it up because joint training does not seem to converge to a good solution. 

\emph{Base Model Training:} The first step in the training pipeline involves training the base model with the temporary output. We describe the training for a single trait for simplicity. The input $x_i$ is an image of size HxWx3 and the label $y_i$ is a 10-dimensional vector representing a discrete distribution over all the votes for trait $t$ and image $x_i$. We adopt this idea from Liu et al \cite{liu2017facial} except our votes are weighted. To compute $y_i$, we must compute its unnormalized form $y\prime_i$ and then normalize. We compute each entry $k$ of $y\prime_i$ as follows:
\begin{equation}
\begin{array}{c}
      y\prime_{ik} = \sum_{j \epsilon \Omega} w_{j} \\
      s.t. \\
      \Omega = \{j | 0.1(k-1) < v_{ij} < 0.1k\}
\end{array}
\end{equation}
And then normalize.
\begin{equation}
    y_{ik} = \frac{y\prime_{ik}}{\sum_p {y\prime_{ip}}}
\end{equation}
The loss is the KL-divergence between the true distribution $y_i$ and the predicted distribution $\bar{o}_i$. This provides the same gradient as cross-entropy but is easier to interpret for the case that the labels $y_i$ are not a 1-hot encoding. We train on 1 pass through the data with the Adam \cite{kingma2014adam} optimizer and an initial learning rate of 1e-4. This prepares the base model for the second phase of training. 

\emph{Voter Model Training:} The second step in the training pipeline involves training the voter model which was described in \ref{sec: architecture}. The goal is the learn the parameters of $\phi$ and the entries of the embedding matrix $E$. The input is now the image $x_i$ and a random voter id $j$, where $j$ comes from the set of voters that have voted on $x_i$, and the label is a 1-hot encoding of the vote $v_{ij}$. We keep the base network frozen allowing us to train with batch size 1000. We use the Adam \cite{kingma2014adam} optimizer with learning rate 1e-3 and the standard cross-entropy loss function. This second step of training is much faster than the first step. 
\subsubsection{Testing}

At test time, the model takes in a test image $x_i$ and returns the output $y_i$ of the aggregator step. We randomly sample a set of 200 voter ids for testing, allowing us to achieve a stable prediction.

\section{Experimental Results \& Discussion}
We implement our Photofeeler-D3 network in Keras \cite{chollet2015keras} on a p2.xlarge AWS instance \cite{aws}. First we evaluate different hyperparameters on our small dataset. Then we evaluate our best set of hyperparameters on the large female and the large male datasets. We evaluate this model against human voters to understand \emph{How many human votes are the model's predictions worth?} Finally, we compare against OAIPs and the architectures in the FBP task. 

\subsection{Hyperparameter Selection}
The key hyperparameters in the Photofeeler-D3 architecture are the image input size, the base network architecture, and the output type. We conduct experiments on the small dataset to confirm the best hyperparameters, and then apply them when training on our large datasets. 

\emph{Image Size:} Standard CNNs for classification use somewhere between a 224x224 and 300x300 image size \cite{vgg16, xception} for their inputs. However we noticed that in images this small, the subject's face is not always clear. Since facial expressions are very important to a voter's impression \cite{pfexpr}, we conjecture that larger image sizes will lead to improved performance. To verify, we conduct an experiment where we fix the base network and the output type, and vary the image size. We show in Table 1 (left) that using the maximum image size of 600x600 achieves almost 9\% higher correlation on the small test set than 224x224. When the images are small the facial expression is not clearly visible, so the model struggles to correctly evaluate the image. 

\begin{table}
\centering
\caption{Quantitative comparison of different hyperparameters for attractiveness prediction on the small dataset. On the left we have image size, in the middle we compare architectures, and on the right we compare output types. This table shows that the best hyperparameters are 600x600 for image size, Xception of architecture, and voter modeling for output type.}
\begin{tabular}{|l|r|} 
\hline
\textbf{Image Size} & \multicolumn{1}{l|}{\textbf{PC (\%)}}  \\ 
\hline
224x224             & 49.3                                   \\ 
\hline
312x312             & 52.0                                   \\ 
\hline
448x448             & 56.8                                   \\ 
\hline
600x600             & \textbf{59.1}                                   \\
\hline
\end{tabular}
\hfill
\begin{tabular}{|l|r|} 
\hline
\textbf{Architecture} & \multicolumn{1}{l|}{\textbf{PC (\%)}}  \\ 
\hline
VGG16 \cite{vgg16} & 42.3                                   \\ 
\hline
ResNet50 \cite{resnet}              & 44.6                                   \\ 
\hline
Inception ResNetV2 \cite{inception4}   & 47.5                                   \\ 
\hline
Xception   \cite{xception}           & \textbf{49.6}                                   \\
\hline
\end{tabular}
\hfill
\begin{tabular}{|l|r|} 
\hline
\textbf{Output Type}  & \multicolumn{1}{l|}{\textbf{PC (\%)}}  \\ 
\hline
Regression            & 39.3                                   \\ 
\hline
Classification        & 47.5                                   \\ 
\hline
Distribution Modeling & 51.3                                   \\ 
\hline
Voter Modeling        & \textbf{54.5}                                   \\
\hline
\end{tabular}
\end{table}

\emph{Architecture:} It's always hard to determine the best base model for a given task, so we tried four standard architectures \cite{inception4, xception, vgg16, resnet} on our task and evaluated them on the small dataset. Table 1 (middle) shows that the Xception \cite{xception} architecture outperforms the others, which is surprising since InceptionResNetV2 \cite{inception4} outperforms Xception on ILSVRC \cite{imagenet}. One explanation is that the Xception architecture should be easier-to-optimize than the InceptionResNetV2. It contains far fewer parameters and a simpler gradient flow \cite{xception}. Since our training dataset is noisy, the gradients will be noisy. When the gradients are noisy, the easier-to-optimize architecture should outperform. 

\emph{Output Type:} There are four main output types to choose from: regression \cite{xu2015new, xu2018transferring}, classification \cite{xu2018crnet, vgg16}, distribution modeling \cite{liu2017facial, fan2018label}, and voter modeling. The results are shown in Table 1 (right). For regression \cite{xu2018transferring} the output is a single neuron that predicts a value in range $[0,1]$, the label is the weighted average of the normalized votes, and the loss is mean squared error (MSE). This performs the worst because the noise in the training set leads to poor gradients which are a large problem for MSE. Classification \cite{xu2018crnet} involves a 10-class softmax output where the labels are a 1-hot encoding of the rounded population mean score. We believe this leads to improved performance because the gradients are smoother for cross-entropy loss. Distribution modeling  \cite{fan2018label, liu2017facial} with weights, as described in section 3.2.2, gives more information to the model. Rather than a single number, it gives a discrete distribution over the votes for the input image. Feeding this added information to the model increases test set correlation by almost 5\%. Finally we note that voter modelling, as described in section 3.2.1, provides another 3.2\% increase. We believe this comes from modeling individual voters rather than the sample mean of what could be very few voters.

We select the hyperparameters with the best performance on the small dataset, and apply them to the large male and female datasets. The results are displayed in Table 2. We notice a large increase in performance from the small dataset because we have 10x more data. However we notice that the model's predictions for attractiveness are consistently poorer than those for trustworthiness and smartness for men, but not for women. This shows that male attractiveness in photos is a more complex/harder-to-model trait. There are a lot of subtleties to what makes a male subject attractive for dating. 

\begin{table}
\centering
\caption{Correlation results of Photofeeler-D3 model on large datasets for both sexes}
\begin{tabular}{|l|r|r|r|} 
\hline
\textbf{Dataset} & \multicolumn{1}{l|}{\textbf{Smart}} & \multicolumn{1}{l|}{\textbf{Trustworthy}} & \multicolumn{1}{l|}{\textbf{Attractive}}  \\ 
\hline
Large Female     & 81.4                                & 83.2                                      & 81.6                                        \\ 
\hline
Large Male       & 80.4                                & 80.6                                      & 74.3                                      \\
\hline
\end{tabular}
\end{table}

\subsection{Photofeeler-D3 vs. Humans}
While Pearson correlation gives a good metric for benchmarking different models, we want to directly compare model predictions to human votes. We devised a test to answer the question: \emph{How many human votes are the model's prediction worth?}. For each example in the test set with over 20 votes, we take the normalized weighted average of all but 15 votes and make it our \emph{truth} score. Then from the remaining 15 votes, we compute the correlation between using 1 vote and the truth score, 2 votes and the truth score, and so on until 15 votes and the truth score. This gives us a correlation curve for up to 15 human votes. We also compute the correlation between the model's prediction and \emph{truth} score. The point on the human correlation curve that matches the correlation of the model gives us the number of votes the model is worth. We do this test using both normalized, weighted votes and raw votes. Table 3 shows that the model is worth an averaged 10.0 raw votes and 4.2 normalized, weighted votes - which means it is better than any single human. Relating it back to online dating, this means that using the Photofeeler-D3 network to select the best photos is as accurate as having 10 people of the opposite sex vote on each image. This means the Photofeeler-D3 network is the first provably reliable OAIP for DPIP. Also this shows that normalizing and weighting the votes based on how a user tends to vote using Photofeeler's algorithm increases the significance of a single vote. As we anticipated, female attractiveness has a significantly higher correlation on the test set than male attractiveness, yet it is worth close to the same number of human votes. This is because male votes on female subject images have a higher correlation with each other than female votes on male subject images. This shows not just that predicting male attractiveness from photos is a more complex task than predicting female attractiveness from photos, but that it is equally more complex for humans as for AI. So even though AI performs worse on the task, humans perform \emph{equally worse} meaning that the ratio stays close to the same. 

\begin{table}
\centering
\caption{Quantitative study showing the number of human votes the model's predictions are worth with respect to each trait. Normalized votes indicates that the votes have gone through Photofeeler's vote weighting and normalizing process. Unnormalized votes contain more noise, therefore the model's predictions are worth more unnormalized votes than normalized ones.}
\begin{tabular}{|l|l|r|r|r|r|} 
\hline
\textbf{Dataset} & \textbf{Norm. Votes?} & \multicolumn{1}{l|}{\textbf{Smart (\#votes)}} & \multicolumn{1}{l|}{\textbf{Trustworthy (\#votes)}} & \multicolumn{1}{l|}{\textbf{Attractive (\#votes)}} & \multicolumn{1}{l|}{\textbf{Mean (\#votes)}}  \\ 
\hline
Large Female     & yes                        & 4.4                                           & 5.0                                                 & 2.7                                                    & 4.0                                           \\ 
\hline
Large Female     & no                         & 11.6                                           & 14.5                                                 & 5.2                                                    & 10.4                                           \\ 
\hline
Large Male       & yes                        & 4.9                                           & 5.2                                                 & 3.1                                                    & 4.4                                          \\ 
\hline
Large Male       & no                         & 11.2                                           & 11.6                                                 & 5.9                                                    & 9.6                                          \\
\hline
\end{tabular}
\end{table}

\subsection{Photofeeler-D3 vs. OAIPs}
To compare to OAIPs, we evaluate prettyscale.com \cite{prettyscale}, hotness.ai \cite{hotness}, and the Photofeeler-D3 network on the London Faces dataset \cite{londonface}. For prettyscale.com and hotness.ai, we use results from an online study \cite{study}. Table 4 shows that our model outperforms both of these by at least 28\% correlation. Photofeeler is the largest online voting platform in the world, therefore the PDD is one of the largest datasets in the world for attractiveness prediction \cite{pfmain}. Through leveraging this data and applying the voter modeling technique, we achieve state-of-the-art performance in OAIPs.

\begin{table}
\centering
\caption{Quantitative comparison of Photofeeler-D3 against other OAIPs on the London Faces Dataset.}
\begin{tabular}{|l|r|} 
\hline
\textbf{OAIP} & \multicolumn{1}{l|}{\textbf{PC (\%)}}  \\ 
\hline
prettyscale.com \cite{prettyscale} & 53                                     \\ 
\hline
hotness.ai \cite{hotness}       & 52                                     \\ 
\hline
Photofeeler-D3             & \textbf{81}                                     \\
\hline
\end{tabular}
\end{table}

\subsection{Photofeeler-D3 in FBP}

In FBP there are 2 main datasets: the SCUT-FBP dataset \cite{xie2015scut} and the HotOrNot dataset \cite{hotornot}. The SCUT-FBP dataset contains 500 female subject images with 10 votes per image from both male and female voters rating the subject's attractiveness from 1-7. The task is to predict the average attractiveness score for an image. This task is different from DPIP for a few reasons: there are only 10 votes - meaning there will be quite a bit of noise; the voters are both male and female, not just male; and the images are not natural, they are neutral faces looking forward into the camera. In the literature, we find some works that only show the best run on the dataset \cite{xu2015new, liu2017facial, xu2018crnet, fan2018label}, and other works that do a 5-fold cross validation \cite{xu2017facial, gaol2018automatic, xu2018transferring} on the dataset. We test our system both ways. We use only the Pearson correlation metric because our scale is from 0-1 whereas the dataset has a scale from 1-7. The Photofeeler-D3 architecture has 3 outputs, one for each trait. To adapt to this dataset, we use only the attractiveness output. All results are shown in Table 5. We show that without any training on the dataset, the Photofeeler-D3 architecture achieves 89\% best run and 78\% in cross validation. Although this is not state-of-the-art, these are still good scores considering how different the task is. If we allow the network to retrain we get 91\% cross validation and 92\% as the best run. This is the best score for cross validation. Additionally, we believe that all of the architectures are getting quite close to the limit on the dataset since there are only 500 examples with 10 votes each. Anything above 90\% correlation is probably fitting the noise of the dataset. We notice that with our dataset, using the average of 10 raw votes is only 87\% correlated with using the average of all the votes. 

The HotOrNot \cite{hotornot} dataset contains 2000 images, 50\% male subjects and 50\% female subjects. Each image has been voted on by over 100 people of the opposite sex. Results are available in Table 5. All other FBP methods \cite{xu2018crnet, gray2010predicting, wang2014attractive, gray2010predicting} first use the Viola-Jones algorithm to crop out the faces and then forward pass their models. Our method takes in the full image, resizes it to 600x600, and forward passes the Photofeeler-D3 network. We show that without any training on this dataset, we achieve 55.9\% cross validation accuracy, outperforming the next best by 7.6\%. Another interesting observation is that our model achieves 68\% correlation with the 1000 females and 42\% correlation with the 1000 males. This reinforces the hypothesis that male attractiveness is a much more complex function to learn than female attractiveness. 

\begin{table}
\centering
\caption{Quantitative Analysis of different models on the Facial Beauty Prediction Task on both the SCUT-FBP dataset and the HotOrNot dataset.}
\begin{tabular}{|l|r|r|l|} 
\hline
\textbf{Architecture} & \multicolumn{1}{l|}{\textbf{SCUT-FBP Best Run}} & \multicolumn{1}{l|}{\textbf{SCUT-FBP 5 Fold CV}} & \textbf{HotOrNot}                   \\ 
\hline
MLP \cite{xu2017facial}                  & 76                                              & 71                                               & -                                   \\ 
\hline
AlexNet-1 \cite{gaol2018automatic}            & 90                                              & 84                                               & -                                   \\ 
\hline
AlexNet-2   \cite{gaol2018automatic}          & 92                                              & 88                                               & -                                   \\ 
\hline
PI-CNN  \cite{xu2017facial}              & 87                                              & 86                                               & -                                   \\ 
\hline
CF        \cite{xu2015new}            & 88                                              & \multicolumn{1}{l|}{-}                           & -                                   \\ 
\hline
LDL       \cite{fan2018label}           & \textbf{93}                                     & \multicolumn{1}{l|}{-}                           & -                                   \\ 
\hline
DRL      \cite{liu2017facial}             & \textbf{93}                                     & \multicolumn{1}{l|}{-}                           & -                                   \\ 
\hline
MT-CNN  \cite{gaol2018automatic}              & 92                                              & 90                                               & -                                   \\ 
\hline
CR-Net   \cite{xu2018crnet}             & 87                                              & \multicolumn{1}{l|}{-}                           & \multicolumn{1}{r|}{48.2}           \\ 
\hline
TRDB   \cite{xu2018transferring}               & 89                                              & 86                                               & \multicolumn{1}{r|}{46.8}           \\ 
\hline
AAE    \cite{wang2014attractive}               & \multicolumn{1}{l|}{-}                          & \multicolumn{1}{l|}{-}                           & \multicolumn{1}{r|}{43.7}           \\ 
\hline
Multi-Scale \cite{gray2010predicting}          & \multicolumn{1}{l|}{-}                          & \multicolumn{1}{l|}{-}                           & \multicolumn{1}{r|}{45.8}           \\ 
\hline
Photofeeler-D3 - No Training    & 89                                              & 78                                               & \multicolumn{1}{r|}{\textbf{58.7}}  \\ 
\hline
Photofeeler-D3 - Retraining     & 92                                              & \textbf{91}                                      & -                                   \\
\hline
\end{tabular}
\end{table}

\section{Conclusion}

In this work we propose the Photofeeler-D3 architecture that, taking advantage of the Photofeeler Dating Dataset and the concept of voter modeling, achieves state-of-the-art results. Additionally, we demonstrate that using our model to select the best dating photos is as accurate as having 10 humans vote on each photo and selecting the best average score. Through this work, we also conclude that Photofeeler's normalizing and weighting algorithm dramatically decreases noise in the votes. Finally we note that although male attractiveness seems to be more difficult to model than female attractiveness, it is equally more difficult for both humans and AI. 

\bibliographystyle{unsrt}
\bibliography{main}
 
  %%% Remove comment to use the external .bib file (using bibtex).


% George Kour and Raid Saabne.
% \newblock Real-time segmentation of on-line handwritten arabic script.
% \newblock In {\em Frontiers in Handwriting Recognition (ICFHR), 2014 14th
%   International Conference on}, pages 417--422. IEEE, 2014.

% George Kour and Raid Saabne.
% \newblock Fast classification of handwritten on-line arabic characters.
% \newblock In {\em Soft Computing and Pattern Recognition (SoCPaR), 2014 6th
%   International Conference of}, pages 312--318. IEEE, 2014.

% \bibitem{hadash2018estimate}
% Guy Hadash, Einat Kermany, Boaz Carmeli, Ofer Lavi, George Kour, and Alon
% \newblock Estimate and replace: A novel approach to integrating deep neural
%   networks with existing applications.
% \newblock {\em arXiv preprint arXiv:1804.09028}, 2018.

% \end{thebibliography}
% \printbibliography
\end{document}